\documentclass[runningheads]{llncs}
\usepackage[nolist]{acronym}
\usepackage{amsmath,graphicx}
\usepackage{algorithm}
\usepackage{algorithmic}
\usepackage{amsmath,amssymb,amsfonts}
\usepackage{caption}
\usepackage{capt-of}
\usepackage{hyperref}
\usepackage[capitalise, noabbrev]{cleveref}
\usepackage{makecell}
\usepackage{mathtools}
\usepackage{multirow}
\usepackage{siunitx}
\DeclareSIUnit\pixel{px}
\let\oldthefigure\thefigure% Capture figure numbering scheme
\let\oldthetable\thetable% Capture table numbering scheme
\newcommand{\supplementaryfigures}{%
  \setcounter{figure}{0}% Reset figure counter
  \setcounter{table}{0}% Reset figure counter
  \renewcommand{\thefigure}{S\oldthefigure}% Prefix figure number with S
  \renewcommand{\thetable}{S\oldthetable}% Prefix figure number with S
}
\begin{document}
\title{Adaptive Region Selection for Active Learning in Whole Slide Image Semantic Segmentation}
% with increased robustness}
%
\titlerunning{Adaptive Region Selection for AL in WSI Semantic Segmentation}
% \author{Anonymous}
% \institute{Anonymous Organization\\
% \email{***@***.***}}
\author{Jingna Qiu\inst{1} \and 
Frauke Wilm\inst{1,2} \and 
Mathias Öttl\inst{2} \and
Maja Schlereth\inst{1} \and
Chang Liu\inst{2} \and
Tobias Heimann\inst{3} \and
Marc Aubreville\inst{4} \and
Katharina Breininger\inst{1}}
% index{Qiu, Jingna, Wilm, Frauke, Öttl, Mathias, Schlereth, Maja, Liu, Chang, Heimann, Tobias, Aubreville, Marc, Breininger, Katharina}
\authorrunning{Qiu et al.}
\institute{Department Artificial Intelligence in Biomedical Engineering, Friedrich-Alexander-Universität Erlangen-Nürnberg, Erlangen, Germany \\
\email{jingna.qiu@fau.de} \and
Pattern Recognition Lab, Department of Computer Science, \\ Friedrich-Alexander-Universität Erlangen-Nürnberg, Erlangen, Germany \and
Digital Technology and Innovation, Siemens Healthineers, Erlangen, Germany \and
Technische Hochschule Ingolstadt, Ingolstadt, Germany\\}
\maketitle              % typeset the header of the contribution
\begin{abstract}
% The abstract should briefly summarize the contents of the paper in
% 15--250 words.
The process of annotating histological gigapixel-sized \acp{wsi} at the pixel level for the purpose of training a supervised segmentation model is time-consuming. Region-based \ac{al} involves training the model on a limited number of annotated image regions instead of requesting annotations of the entire images. These annotation regions are iteratively selected, with the goal of optimizing model performance while minimizing the annotated area. The standard method for region selection evaluates the informativeness of all square regions of a specified size and then selects a specific quantity of the most informative regions. We find that the efficiency of this method highly depends on the choice of \ac{al} step size (i.e., the combination of region size and the number of selected regions per \ac{wsi}), and a suboptimal \ac{al} step size can result in redundant annotation requests or inflated computation costs. This paper introduces a novel technique for selecting annotation regions adaptively, mitigating the reliance on this \ac{al} hyperparameter.  Specifically, we dynamically determine each region by first identifying an informative area and then detecting its optimal bounding box, as opposed to selecting regions of a uniform predefined shape and size as in the standard method. We evaluate our method using the task of breast cancer metastases segmentation on the public CAMELYON16 dataset and show that it consistently achieves higher sampling efficiency than the standard method across various \ac{al} step sizes. With only $2.6\%$ of tissue area annotated, we achieve full annotation performance and thereby substantially reduce the costs of annotating a \ac{wsi} dataset. The source code is available at \url{https://github.com/DeepMicroscopy/AdaptiveRegionSelection}
\end{abstract}
\section{Introduction}
\acresetall
Semantic segmentation on histological \acp{wsi} allows precise detection of tumor boundaries, thereby facilitating the assessment of metastases~\cite{guo2019fast} and other related analytical procedures~\cite{wilm2022pan}. However, pixel-level annotations of gigapixel-sized \acp{wsi} (e.g. $100,000 \times 100,000$ \si{pixels}) for training a segmentation model are difficult to acquire. For instance, in the CAMELYON16 breast cancer metastases dataset~\cite{litjens20181399}, $49.5\%$ of \acp{wsi} contain metastases that are smaller than $1\%$ of the tissue, requiring a high level of expertise and long inspection time to ensure exhaustive tumor localization; whereas other \acp{wsi} have large tumor lesions and require a substantial amount of annotation time for boundary delineation~\cite{xu2022clinical}. Identifying potentially informative image regions (i.e., providing useful information for model training) allows requesting the minimum amount of annotations for model optimization, and a decrease in annotated area reduces both localization and delineation workloads. The challenge is to effectively select annotation regions in order to achieve full annotation performance with the least annotated area, resulting in high sampling efficiency.

We use region-based \ac{al}~\cite{mackowiak2018cereals} to progressively identify annotation regions, based on iteratively updated segmentation models. Each region selection process consists of two steps. First, the prediction of the most recently trained segmentation model is converted to a priority map that reflects informativeness of each pixel. Existing studies on \acp{wsi} made extensive use of informativeness measures that quantify model uncertainty (e.g., least confidence~\cite{lai2021joint}, maximum entropy~\cite{jin2021reducing} and highest disagreement between a set of models~\cite{yang2017suggestive}). The enhancement of priority maps, such as highlighting easy-to-label pixels~\cite{mackowiak2018cereals}, edge pixels~\cite{kasarla2019region} or pixels with a low estimated segmentation quality~\cite{colling2020metabox+}, is also a popular area of research. Second, on the priority map, regions are selected according to a region selection method. Prior works have rarely looked into region selection methods; the majority followed the standard approach~\cite{mackowiak2018cereals} where a sliding window divides the priority map into fixed-sized square regions, the selection priority of each region is calculated as the cumulative informativeness of its constituent pixels, and a number of regions with the highest priorities are then selected. In some other works, only non-overlapping or sparsely overlapped regions were considered to be candidates~\cite{lai2021joint,yang2017suggestive}. Following that, some works used additional criteria to filter the selected regions, such as finding a representative subset~\cite{jin2021reducing,yang2017suggestive}. All of these works selected square regions of a manually predefined size, disregarding the actual shape and size of informative areas.

This work focuses on region selection methods, a topic that has been largely neglected in literature until now, but which we show to have a great impact on \ac{al} sampling efficiency (i.e., the annotated area required to reach the full annotation performance). We discover that the sampling efficiency of the aforementioned standard method decreases as the \ac{al} step size (i.e., the annotated area at each \ac{al} cycle, determined by the multiplication of the region size and the number of selected regions per \ac{wsi}) increases. To avoid extensive \ac{al} step size tuning, we propose an adaptive region selection method with reduced reliance on this \ac{al} hyperparameter. Specifically, our method dynamically determines an annotation region by first identifying an informative area with connected component detection and then detecting its bounding box. We test our method using a breast cancer metastases segmentation task on the public CAMELYON16 dataset and demonstrate that determining the selected regions individually provides greater flexibility and efficiency than selecting regions with a uniform predefined shape and size, given the variability in histological tissue structures. Results show that our method consistently outperforms the standard method by providing a higher sampling efficiency, while also being more robust to \ac{al} step size choices. Additionally, our method is especially beneficial for settings where a large \ac{al} step size is desirable due to annotator availability or computational restrictions.

\section{Method}
\subsection{Region-based Active Learning for \ac{wsi} Annotation}
\label{sec:method_region_based_active_learning}
We are given an unlabeled pool $\mathcal{U}=\{X_{1}\dots X_{n}\}$, where $X_{i}\in \mathbb{R}^{W_{i}\times H_{i}}$ denotes the $i^{th}$ \ac{wsi} with width $W_{i}$ and height $H_{i}$. Initially, $X_{i}$ has no annotation; regions are iteratively selected from it and annotated across \ac{al} cycles. We denote the $j^{th}$ annotated rectangular region in $X_{i}$ as $R_{ij}=(c_x^{ij}, c_y^{ij}, w^{ij}, h^{ij})$, where ($c_x^{ij}, c_y^{ij}$) are the center coordinates of the region and $w^{ij}, h^{ij}$ are the width and height of that region, respectively. In the standard region selection method, where fixed-size square regions are selected, $w^{ij}=h^{ij}=l, \forall i, j$, where $l$ is predefined. 
\begin{figure}[h]
\begin{minipage}[b]{1.0\linewidth}
  \centering  \centerline{\includegraphics[width=\textwidth]{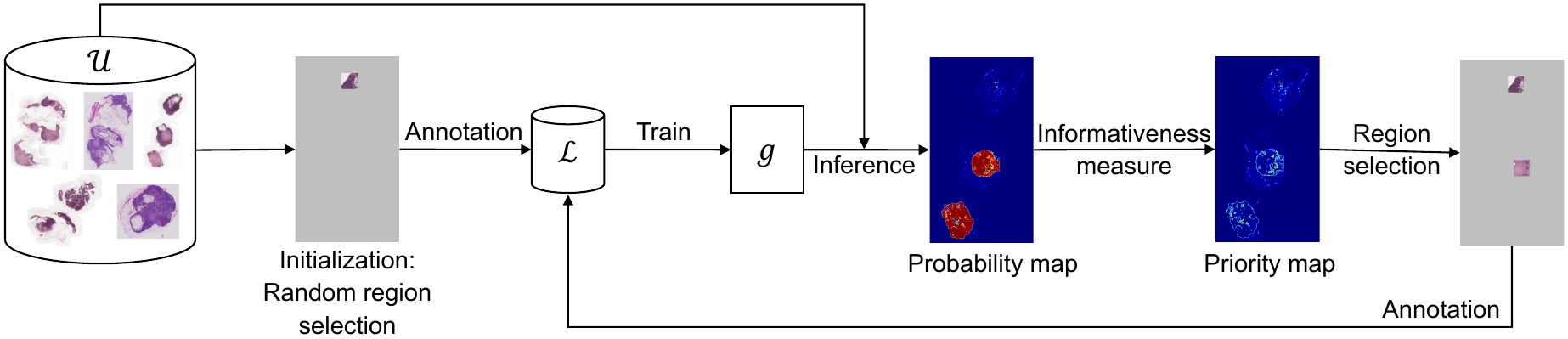}}
\end{minipage}
\caption{Region-based \ac{al} workflow for selecting annotation regions. The exemplary selected regions are of size $8192 \times 8192$ \si{pixels}. (Image resolution: $0.25$~\si{\micro\meter\per\pixel})} 
\label{fig:workflow}
\end{figure}

Fig.~\ref{fig:workflow} illustrates the workflow of region-based \ac{al} for \ac{wsi} annotation. The goal is to iteratively select and annotate potentially informative regions from \acp{wsi} in $\mathcal{U}$ to enrich the labeled set $\mathcal{L}$ in order to effectively update the model $g$. To begin, $k$ regions (each containing at least 10\% of tissue) per \ac{wsi} are randomly selected and annotated to generate the initial labeled set $\mathcal{L}$. The model $g$ is then trained on $\mathcal{L}$ and predicts on $\mathcal{U}$ to select $k$ new regions from each \ac{wsi} for the new round of annotation. The newly annotated regions are added to $\mathcal{L}$ for retraining $g$ in the next \ac{al} cycle. The train-select-annotate process is repeated until a certain performance of $g$ or annotation budget is reached. 

The selection of $k$ new regions from $X_{i}$ is performed in two steps based on the model prediction $P_{i}=g(X_{i})$. First, $P_{i}$ is converted to a priority map $M_{i}$ using a per-pixel informativeness measure. Second, $k$ regions are selected based on $M_{i}$ using a region selection method. The informativeness measure is not the focus of this study, we therefore adopt the most commonly used one that quantifies model uncertainty (details in Section~\ref{ssec:implementation}). Next we describe the four region selection methods evaluated in this work.

\subsection{Region Selection Methods}
\label{sec:method_region_based_active_learning}
\begin{figure}
    \centering
    \begin{minipage}{.35\textwidth}
        \centering
        \includegraphics[width=0.8\linewidth]{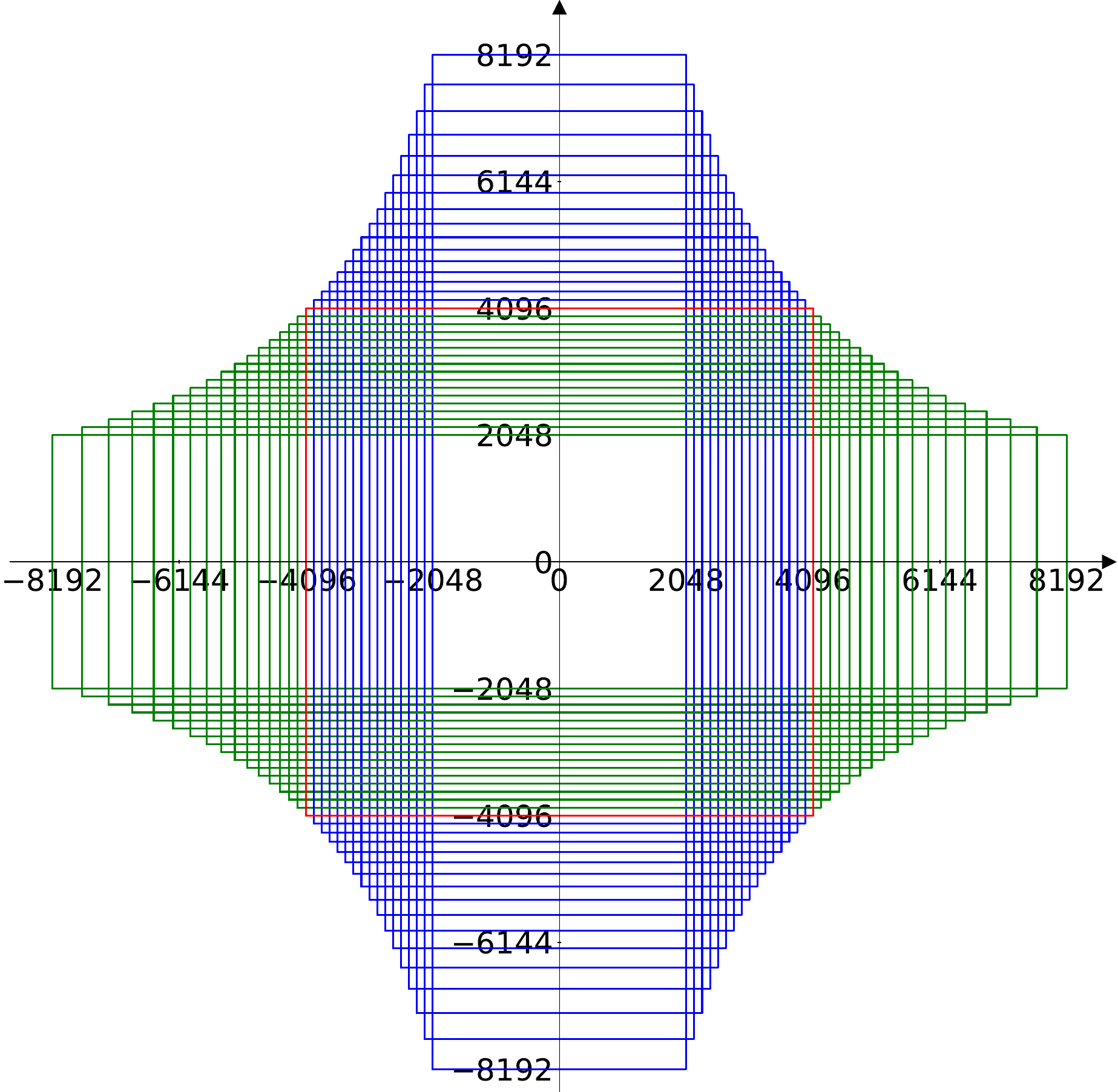}
    \end{minipage}%
    \begin{minipage}{0.65\textwidth}
        \centering
        \includegraphics[width=\linewidth]{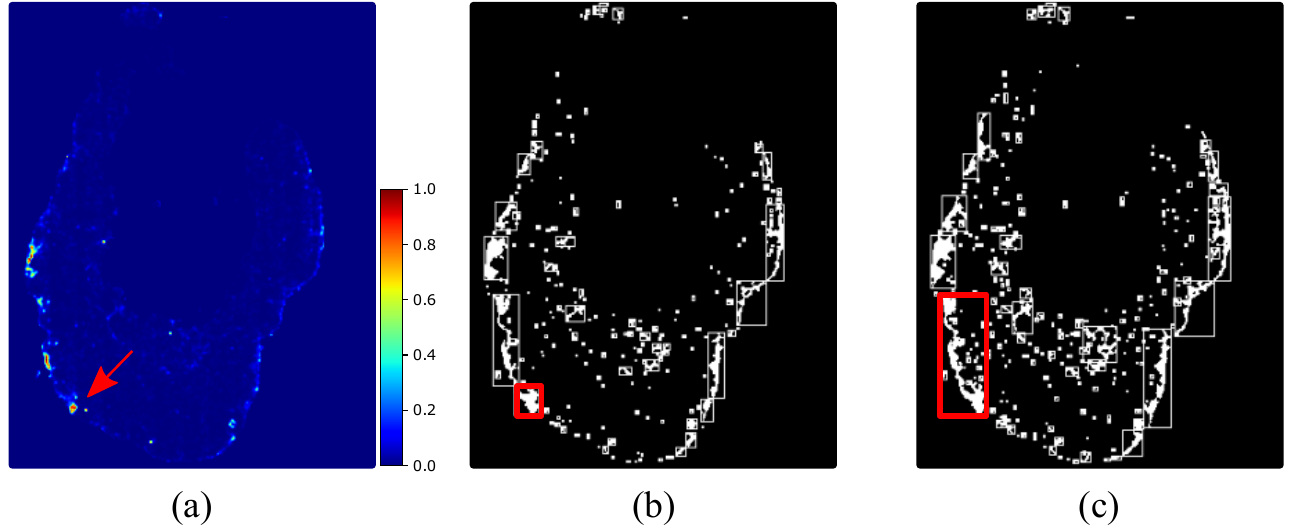}
    \end{minipage}
    \begin{minipage}{.35\textwidth}
        \centering
        \captionsetup{width=0.9\linewidth}
        \caption{\textit{Standard (non-square)}: Region candidates for $l=\SI{8192}{pixels}$. }
        \label{fig:non-square_region_candidates}
    \end{minipage}%
    \begin{minipage}{0.65\textwidth}
        \centering
        \captionsetup{width=\linewidth}
        \caption{\textit{Adaptive}: (a) Priority map $M_{i}$ and the highest priority pixel (arrow). (b-c) Bisection search of $\tau$: (b) $\tau=99^{th}$, (c) $\tau=98.5^{th}$.}
        \label{fig:adaptive_method_illustration}
    \end{minipage}
\end{figure}

\noindent \textit{\textbf{Random}} This is the baseline method where $k$ regions of size $l\times l$ are randomly selected. Each region contains at least $10\%$ of tissue and does not overlap with other regions.  \textit{\textbf{Standard}}~\cite{mackowiak2018cereals} $M_i$ is divided into overlapping regions of a fixed size $l\times l$ using a sliding window with a stride of $\SI{1}{pixel}$. The selection priority of each region is calculated as the summed priority of the constituent pixels, and $k$ regions with the highest priorities are then selected. Non-maximum suppression is used to avoid selecting overlapping regions. \textit{\textbf{Standard (non-square)}} We implement a generalized version of the standard method that allows non-square region selections by including multiple region candidates centered at each pixel with various aspect ratios. To save computation and prevent extreme shapes, such as those with a width or height of only a few pixels, we specify a set of candidates as depicted in Fig.~\ref{fig:non-square_region_candidates}. Specifically, we define a variable region width $w$ as spanning from $\frac{1}{2}l$ to $l$ with a stride of \SI{256}{pixels} and determine the corresponding region height as $h=\frac{l^2}{w}$. \textit{\textbf{Adaptive}} (proposed) Our method allows for selecting regions with variable aspect ratios and sizes to accommodate histological tissue variability. The $k$ regions are selected sequentially; when selecting the $j^{th}$ region $R_{ij}$ in $X_{i}$, we first set the priorities of all pixels in previously selected regions (if any) to zero. We then find the highest priority pixel $(c_x^{ij}, c_y^{ij})$ on $M_{i}$; a median filter with a kernel size of 3 is applied beforehand to remove outliers. Afterwards, we create a mask on $M_{i}$ with an intensity threshold of $\tau^{th}$ percentile of intensities in $M_{i}$, detect the connected component containing $(c_x^{ij}, c_y^{ij})$, and select its bounding box. As depicted in Fig.~\ref{fig:adaptive_method_illustration}, $\tau$ is determined by performing a bisection search over $[98, 100]^{th}$ percentiles, such that the bounding box size is in range $[\frac{1}{2}l\times \frac{1}{2}l, \frac{3}{2}l\times \frac{3}{2}l]$. This size range is chosen to be comparable to the other three methods, which select regions of size $l^{2}$. Note that \textit{Standard (non-square)} can be understood as an ablation study of the proposed method \textit{Adaptive} to examine the effect of variable region shape by maintaining constant region size.

\subsection{\Ac{wsi} Semantic Segmentation Framework}
\label{ssec:WSI_segmentation}
This section describes the breast cancer metastases segmentation task we use for evaluating the \ac{al} region selection methods. The task is performed with patch-wise classification, where the \ac{wsi} is partitioned into patches, each patch is classified as to whether it contains metastases, and the results are assembled. \textbf{Training.} The patch classification model $h(\mathbf{x}, \mathbf{w}):\mathbb{R}^{d\times d} \xrightarrow[]{}[0, 1]$ takes as input a patch $\mathbf{x}$ and outputs the probability $p(y=1|\mathbf{x}, \mathbf{w})$ of containing metastases, where $\mathbf{w}$ denotes model parameters. Patches are extracted from the annotated regions at $40\times$ magnification ($0.25$~\si{\micro\meter\per\pixel}) with $d=256$~\si{pixels}. Following~\cite{liu2017detecting}, a patch is labeled as positive if the center $128 \times 128$ \si{pixels} area contains at least one metastasis pixel and negative otherwise. In each training epoch, $20$ patches per \ac{wsi} are extracted at random positions within the annotated area; for \acp{wsi} containing annotated metastases, positive and negative patches are extracted with equal probability. A patch with less than $1\%$ tissue content is discarded. Data augmentation includes random flip, random rotation, and stain augmentation~\cite{macenko2009method}.
\textbf{Inference.} $X_{i}$ is divided into a grid of uniformly spaced patches ($40\times$ magnification, $d=\SI{256}{pixels}$) with a stride $s$. The patches are predicted using the trained patch classification model and the results are stitched to a probability map $P_{i}\in [0, 1]^{W_{i}'\times H_{i}'}$, where each pixel represents a patch prediction. The patch extraction stride $s$ determines the size of $P_{i}$ ($W_{i}'=\frac{W_{i}}{s}, H_{i}'=\frac{H_{i}}{s}$). 

\section{Experiments}
\subsection{Dataset} We used the publicly available CAMELYON16 Challenge dataset~\cite{litjens20181399}, licensed under the Creative Commons CC0 license. The collection of the data was approved by the responsible ethics committee (Commissie Mensgebonden Onderzoek regio Arnhem-Nijmegen). The CAMELYON16 dataset consists of $399$ \ac{he}-stained \acp{wsi} of sentinel axillary lymph node sections. The training set contains $111$ \acp{wsi} with and $159$ \acp{wsi} without breast cancer metastases, and each \ac{wsi} with metastases is accompanied by pixel-level contour annotations delineating the boundaries of the metastases. We randomly split a stratified $30\%$ subset of the training set as the validation set for model selection. The test set contains $48$ \acp{wsi} with and $80$ \acp{wsi} without metastases~\footnote[1]{Test\_114 is excluded due to non-exhaustive annotation, as stated by data provider.}. 

\subsection{Implementation Details}
\label{ssec:implementation}

\noindent \textbf{Training 
Schedules} We use MobileNet\_v2~\cite{sandler2018mobilenetv2} initialized with ImageNet~\cite{russakovsky2015imagenet} weights as the backbone of the patch classification model. It is extended with two fully-connected layers with sizes of $512$ and $2$, followed by a softmax activation layer. The model is trained for up to $500$ epochs using cross-entropy loss and the Adam optimizer~\cite{kingma2014adam}, and is stopped early if the validation loss stagnates for $100$ consecutive epochs. Model selection is guided by the lowest validation loss. The learning rate is scheduled by the one cycle policy~\cite{smith2018disciplined} with a maximum of $0.0005$. The batch size is $32$. We used Fastai v1~\cite{howard2020fastai} for model training and testing. The running time of one \ac{al} cycle (select-train-test) on a single NVIDIA Geforce RTX3080 GPU (10GB) is around 7 hours. \\

\noindent \textbf{Active Learning Setups} Since the CAMELYON16 dataset is fully annotated, we perform \ac{al} by assuming all \acp{wsi} are unannotated and revealing the annotation of a region only after it is selected during the \ac{al} procedure. We divide the \acp{wsi} in $\mathcal{U}$ randomly into five stratified subsets of equal size and use them sequentially. In particular, regions are selected from \acp{wsi} in the first subset at the first \ac{al} cycle, from \acp{wsi} in the second subset at the second \ac{al} cycle, and so on. This is done because \ac{wsi} inference is computationally expensive due to the large patch amount, reducing the number of predicted \acp{wsi} to one fifth helps to speed up \ac{al} cycles. We use an informativeness measure that prioritizes pixels with a predicted probability close to $0.5$ (i.e., $M_{i}=1-2|P_{i}-0.5|$), following~\cite{lewis1995sequential}. We annotate validation \acp{wsi} in the same way as the training \acp{wsi} via \ac{al}.\\ 

\noindent \textbf{Evaluations}
We use the CAMELYON16 challenge metric \ac{froc} score~\cite{bejnordi2017diagnostic} to validate the segmentation framework. To evaluate the \ac{wsi} segmentation performance directly, we use \ac{miou}. For comparison, we follow~\cite{guo2019fast} to use a threshold of $0.5$ to generate the binary segmentation map and report \ac{miou} (Tumor), which is the average \ac{miou} over the $48$ test \acp{wsi} with metastases. We evaluate the model trained at each \ac{al} cycle to track performance change across the \ac{al} procedure.

\subsection{Results} 
\subsubsection{Full Annotation Performance} 
To validate our segmentation framework, we first train on the fully-annotated data (average performance of five repetitions reported). With a patch extraction stride $s=\SI{256}{pixels}$, our framework yields an \ac{froc} score of $0.760$ that is equivalent to the Challenge top $2$, and an \ac{miou} (Tumor) of $0.749$, which is higher than the most comparable method in \cite{guo2019fast} that achieved $0.741$ with $s=\SI{128}{pixels}$. With our framework, reducing $s$ to $\SI{128}{pixels}$ improves both metastases identification and segmentation (\ac{froc} score: $0.779$, \ac{miou} (Tumor): $0.758$). However, halving $s$ results in a $4$-fold increase in inference time. This makes an \ac{al} experiment, which involves multiple rounds of \ac{wsi} inference, extremely costly. Therefore, we use $s=\SI{256}{pixels}$ for all following \ac{al} experiments to compromise between performance and computation costs. Because \acp{wsi} without metastases do not require pixel-level annotation, we exclude the $159$ training and validation \acp{wsi} without metastases from all following \ac{al} experiments. This reduction leads to a slight decrease of full annotation performance (\ac{miou} (Tumor) from $0.749$ to $0.722$).
% and we are more interested in the relative changes in model performance across \ac{al} cycles than the absolute results.

\subsubsection{Comparison of Region Selection Methods}
\begin{figure}[t]
\centering
\begin{minipage}[b]{.9\linewidth}
  \centering  \centerline{\includegraphics[width=\textwidth]{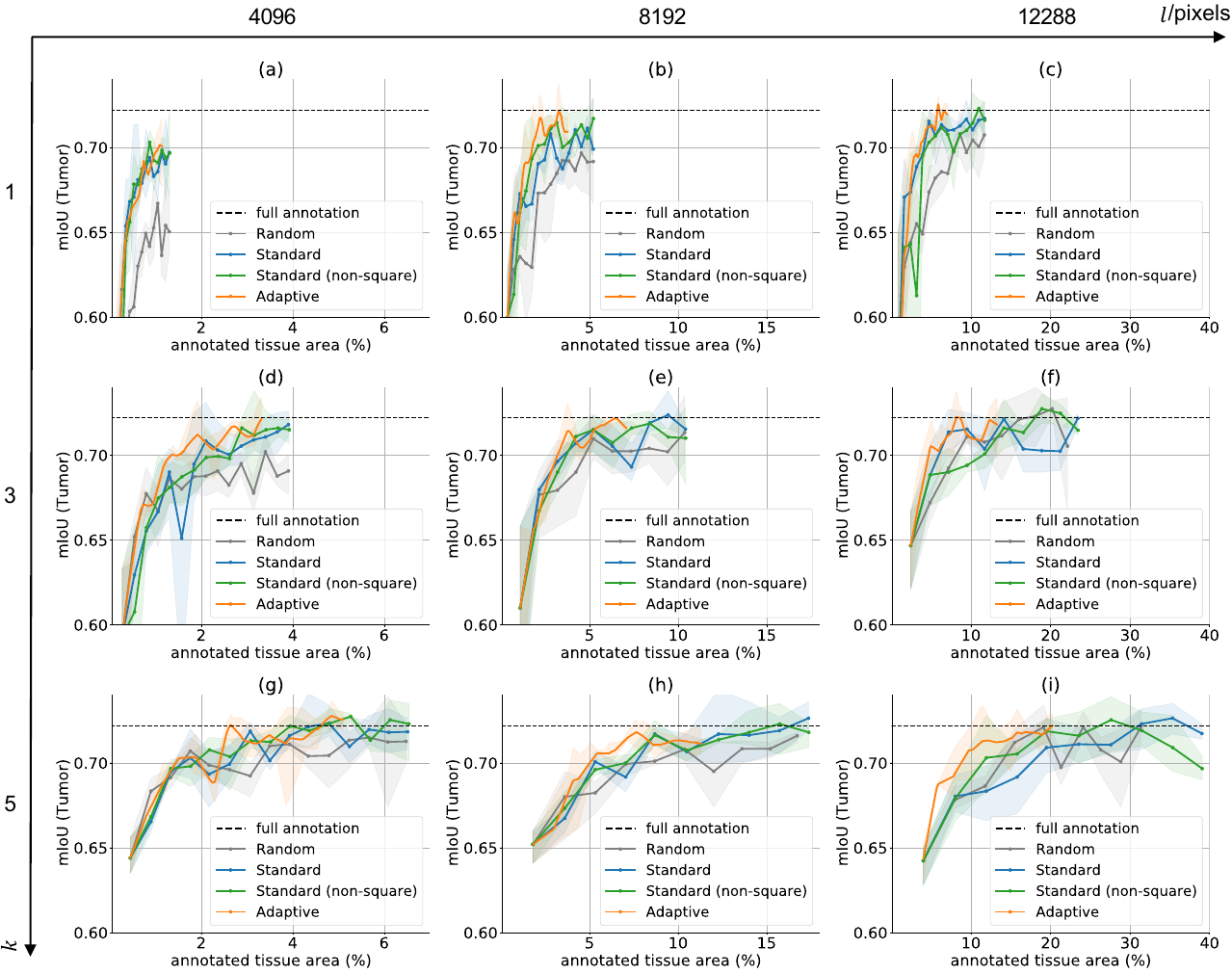}}
\end{minipage}
\caption{\ac{miou} (Tumor) as a function of annotated tissue area (\%) for four region selection methods across various \ac{al} step sizes. Results show average and min/max (shaded) performance over three repetitions with distinct initial labeled sets. The final annotated tissue area of \textit{Random} can be less than \textit{Standard} as it stops sampling a \ac{wsi} if no region contains more than $10\%$ of tissue. Curves of \textit{Adaptive} are interpolated as the annotated area differs between repetitions.}
\label{fig:step_size_result}
\end{figure}

\begin{table}
\centering
\caption{Annotated tissue area (\%) required to achieve full annotation performance. The symbol ``/'' indicates that the full annotation performance is not achieved in the corresponding experimental setting in Fig.~\ref{fig:step_size_result}.}\label{tab1}
\begin{tabular}{m{3.3cm}|m{0.8cm}|m{0.8cm}|m{0.8cm}|m{0.8cm}|m{0.8cm}|m{0.8cm}|m{0.8cm}|m{0.8cm}|m{0.8cm}}
\hline
$k$ & \multicolumn{3}{c|}{1} & \multicolumn{3}{c|}{3} & \multicolumn{3}{c}{5}\\
\hline
$l/\si{pixels}$ & 4096 & 8192 & 12288 & 4096 & 8192 & 12288 & 4096 & 8192 & 12288\\
\Xhline{3\arrayrulewidth}
\textit{Random} & / & / & / & / & / & 18.1 & / & / & /\\
\hline
\textit{Standard} & / & / & / & / & 9.4 & 14.2 & 4.3 & 17.4 & 31.5\\
\hline
\textit{Standard (non-square)} & / & / & 11.0 & / & / & 18.9 & 3.9 & 15.7 & 27.6\\
\hline
\textit{Adaptive} & / & 3.3 & 5.8 & 3.3 & 6.3 & 8.1 & 2.6 & / & 20.0\\
\hline
\end{tabular}
\end{table}

Fig.~\ref{fig:step_size_result} compares the sampling efficiency of the four region selection methods across various \ac{al} step sizes (i.e., the combinations of region size $l\in \{4096, 8192, 12288\}$ \si{pixels} and the number of selected regions per \ac{wsi} $k\in \{1, 3, 5\}$). Experiments with large \ac{al} step sizes perform  $10$ \ac{al} cycles (Fig.~\ref{fig:step_size_result} (e), (f), (h) and (i)); others perform $15$ \ac{al} cycles. All experiments (except for \textit{Random}) use uncertainty sampling. 

\begin{figure}[t]
\centering
\begin{minipage}{\linewidth}
  \centering  
  \centerline{\includegraphics[width=\textwidth]{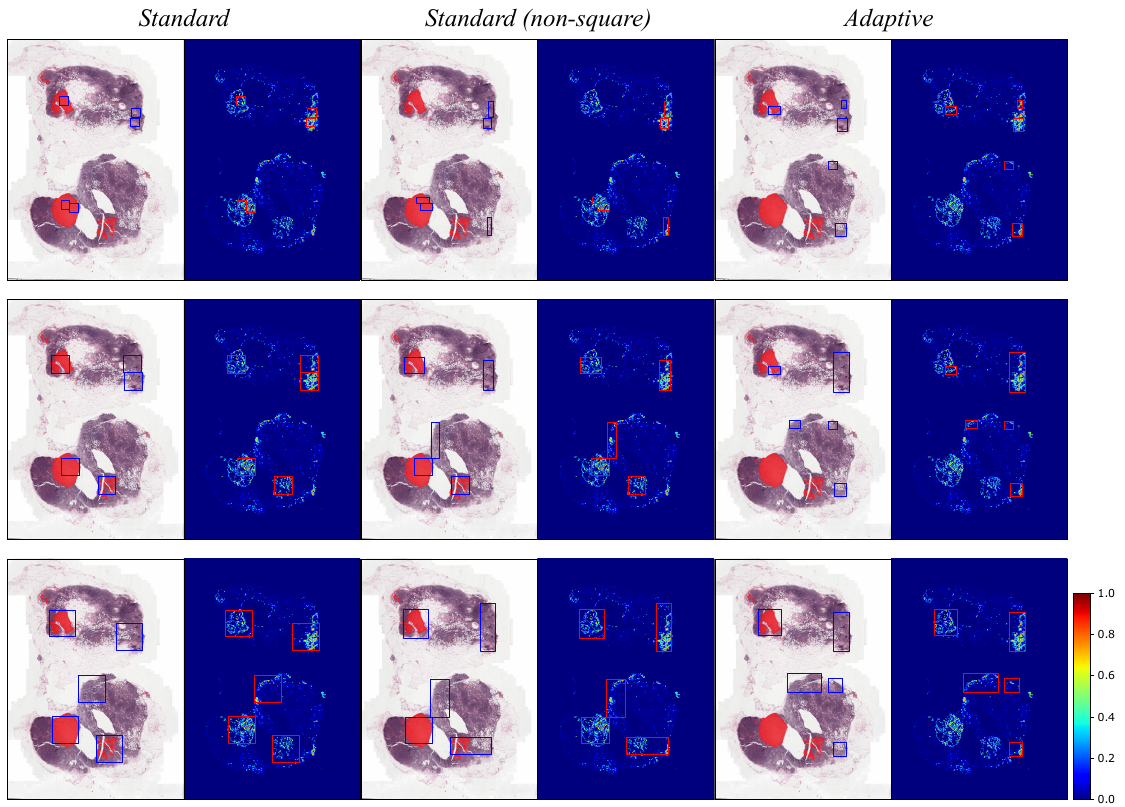}}
\end{minipage}
\caption{Visualization of five regions selected with three region selection methods, applied to an exemplary priority map produced in a second \ac{al} cycle (regions were randomly selected in the first \ac{al} cycle, $k=5, l=4096$ \si{pixels}). Region sizes increase from top to bottom: $l\in\{4096, 8192, 12288\}$ \si{pixels}. Fully-annotated tumor metastases overlaid with \ac{wsi} in red.}
\label{fig:qualitative_result}
\end{figure}

When using region selection method \textit{Standard}, the sampling efficiency advantage of uncertainty sampling over random sampling decreases as \ac{al} step size increases. A small \ac{al} step size minimizes the annotated tissue area for a certain high level of model performance, such as an \ac{miou} (Tumor) of $0.7$, yet requires a large number of \ac{al} cycles to achieve full annotation performance (Fig.~\ref{fig:step_size_result} (a-d)), resulting in high computation costs. A large \ac{al} step size allows for full annotation performance to be achieved in a small number of \ac{al} cycles, but at the expense of rapidly expanding the annotated tissue area (Fig.~\ref{fig:step_size_result} (e), (f), (h) and (i)). Enabling selected regions to have variable aspect ratios does not substantially improve the sampling efficiency, with \textit{Standard (non-square)} outperforming \textit{Standard} only when the \ac{al} step size is excessively large (Fig.~\ref{fig:step_size_result} (i)). However, allowing regions to be of variable size consistently improves sampling efficiency. Table~\ref{tab1} shows that \textit{Adaptive} achieves full annotation performance with fewer \ac{al} cycles than \textit{Standard} for small \ac{al} step sizes and less annotated tissue area for large \ac{al} step sizes. As a result, when region selection method \textit{Adaptive} is used, uncertainty sampling consistently outperforms random sampling. Furthermore, Fig.~\ref{fig:step_size_result} (e-i)) shows that \textit{Adaptive} effectively prevents the rapid expansion of annotated tissue area as \ac{al} step size increases, demonstrating greater robustness to \ac{al} step size choices than \textit{Standard}. This is advantageous because extensive \ac{al} step size tuning to balance the annotation and computation costs can be avoided. This behavior can also be desirable in cases where frequent interaction with annotators is not possible or to reduce computation costs, because the proposed method is more tolerant to a large \ac{al} step size. 

We note in Fig.~\ref{fig:step_size_result} (h) that the full annotation performance is not achieved with \textit{Adaptive} within 15 \ac{al} cycles; in Fig.~\ref{fig:step_size_result_oversample} in the supplementary materials we show that allowing for oversampling of previously selected regions can be a solution to this problem. Additionally, we visualize examples of selected regions in Fig.~\ref{fig:qualitative_result} and show that \textit{Adaptive} avoids two region selection issues of \textit{Standard}: small, isolated informative areas are missed, and irrelevant pixels are selected due to the region shape and size restrictions. 

\section{Discussion and Conclusion}
We presented a new \ac{al} region selection method to select annotation regions on \acp{wsi}. In contrast to the standard method that selects regions with predetermined shape and size, our method takes into account the intrinsic variability of histological tissue and dynamically determines the shape and size for each selected region. Experiments showed that it outperforms the standard method in terms of both sampling efficiency and the robustness to \ac{al} hyperparameters. Although the uncertainty map was used to demonstrate the efficacy of our approach, it can be seamlessly applied to any priority maps. A limitation of this study is that the annotation cost is estimated only based on the annotated area, while annotation effort may vary when annotating regions of equal size. Future work will involve the development of a \ac{wsi} dataset with comprehensive documentation of annotation time to evaluate the proposed method and an investigation of potential combination with self-supervised learning. 

{\textbf{Acknowledgments.} We thank Yixing Huang and Zhaoya Pan (FAU) for their feedback on the manuscript. We gratefully acknowledge support by d.hip campus - Bavarian aim (J.Q. and K.B.) as well as the scientific support and HPC resources provided by the Erlangen National High Performance Computing Center (NHR@FAU) of the Friedrich-Alexander-Universität Erlangen-Nürnberg (FAU). The hardware is funded by the German Research Foundation (DFG).}

\begin{acronym}
\acro{wsi}[WSI]{whole slide image}
\acro{he}[H\&E]{Hematoxylin \& Eosin}
\acro{froc}[FROC]{Free Response Operating Characteristic}
\acro{iou}[IoU]{intersection over union}
\acro{miou}[mIoU]{mean intersection over union}
\acro{al}[AL]{active learning}
\end{acronym}
\bibliographystyle{splncs04}
\bibliography{paper1681}

\begin{thebibliography}{10}
\providecommand{\url}[1]{\texttt{#1}}
\providecommand{\urlprefix}{URL }
\providecommand{\doi}[1]{https://doi.org/#1}

\bibitem{bejnordi2017diagnostic}
Bejnordi, B.E., Veta, M., Van~Diest, P.J., Van~Ginneken, B., Karssemeijer, N.,
  Litjens, G., Van Der~Laak, J.A., Hermsen, M., Manson, Q.F., Balkenhol, M.,
  et~al.: Diagnostic assessment of deep learning algorithms for detection of
  lymph node metastases in women with breast cancer. Jama  \textbf{318}(22),
  2199--2210 (2017)

\bibitem{colling2020metabox+}
Colling, P., Roese-Koerner, L., Gottschalk, H., Rottmann, M.: Metabox+: A new
  region based active learning method for semantic segmentation using priority
  maps. arXiv preprint arXiv:2010.01884  (2020)

\bibitem{guo2019fast}
Guo, Z., Liu, H., Ni, H., Wang, X., Su, M., Guo, W., Wang, K., Jiang, T., Qian,
  Y.: A fast and refined cancer regions segmentation framework in whole-slide
  breast pathological images. Scientific reports  \textbf{9}(1),  1--10 (2019)

\bibitem{howard2020fastai}
Howard, J., Gugger, S.: Fastai: a layered api for deep learning. Information
  \textbf{11}(2), ~108 (2020)

\bibitem{jin2021reducing}
Jin, X., An, H., Wang, J., Wen, K., Wu, Z.: Reducing the annotation cost of
  whole slide histology images using active learning. In: 2021 3rd
  International Conference on Image Processing and Machine Vision (IPMV). pp.
  47--52 (2021)

\bibitem{kasarla2019region}
Kasarla, T., Nagendar, G., Hegde, G.M., Balasubramanian, V., Jawahar, C.:
  Region-based active learning for efficient labeling in semantic segmentation.
  In: 2019 IEEE Winter Conference on Applications of Computer Vision (WACV).
  pp. 1109--1117. IEEE (2019)

\bibitem{kingma2014adam}
Kingma, D.P., Ba, J.: Adam: A method for stochastic optimization. arXiv
  preprint arXiv:1412.6980  (2014)

\bibitem{lai2021joint}
Lai, Z., Wang, C., Oliveira, L.C., Dugger, B.N., Cheung, S.C., Chuah, C.N.:
  Joint semi-supervised and active learning for segmentation of gigapixel
  pathology images with cost-effective labeling. In: Proceedings of the
  IEEE/CVF International Conference on Computer Vision. pp. 591--600 (2021)

\bibitem{lewis1995sequential}
Lewis, D.D., Gale, W.A.: A sequential algorithm for training text classifiers.
  In: Proceedings of the 17th Annual International ACM SIGIR Conference on
  Research and Development in Information Retrieval. p. 3–12. SIGIR '94,
  Springer-Verlag, Berlin, Heidelberg (1994)

\bibitem{litjens20181399}
Litjens, G., Bandi, P., Ehteshami~Bejnordi, B., Geessink, O., Balkenhol, M.,
  Bult, P., et~al.: 1399 h\&e-stained sentinel lymph node sections of breast
  cancer patients: the camelyon dataset. GigaScience  \textbf{7}(6),  giy065
  (2018), data downloaded from the GigaScience database
  \url{http://gigadb.org/dataset/100439}.

\bibitem{liu2017detecting}
Liu, Y., Gadepalli, K., Norouzi, M., Dahl, G.E., Kohlberger, T., Boyko, A.,
  Venugopalan, S., Timofeev, A., Nelson, P.Q., Corrado, G.S., et~al.: Detecting
  cancer metastases on gigapixel pathology images. arXiv preprint
  arXiv:1703.02442  (2017)

\bibitem{macenko2009method}
Macenko, M., Niethammer, M., Marron, J.S., et~al.: A method for normalizing
  histology slides for quantitative analysis. In: 2009 IEEE international
  symposium on biomedical imaging: from nano to macro. pp. 1107--1110. IEEE
  (2009)

\bibitem{mackowiak2018cereals}
Mackowiak, R., Lenz, P., Ghori, O., et~al.: Cereals-cost-effective region-based
  active learning for semantic segmentation. arXiv preprint arXiv:1810.09726
  (2018)

\bibitem{russakovsky2015imagenet}
Russakovsky, O., Deng, J., Su, H., Krause, J., Satheesh, S., Ma, S., Huang, Z.,
  Karpathy, A., Khosla, A., Bernstein, M., et~al.: Imagenet large scale visual
  recognition challenge. International journal of computer vision
  \textbf{115}(3),  211--252 (2015)

\bibitem{sandler2018mobilenetv2}
Sandler, M., Howard, A., Zhu, M., Zhmoginov, A., Chen, L.C.: Mobilenetv2:
  Inverted residuals and linear bottlenecks. In: Proceedings of the IEEE
  conference on computer vision and pattern recognition. pp. 4510--4520 (2018)

\bibitem{smith2018disciplined}
Smith, L.N.: A disciplined approach to neural network hyper-parameters: Part
  1--learning rate, batch size, momentum, and weight decay. arXiv preprint
  arXiv:1803.09820  (2018)

\bibitem{wilm2022pan}
Wilm, F., Fragoso, M., Marzahl, C., Qiu, J., Puget, C., Diehl, L., Bertram,
  C.A., Klopfleisch, R., Maier, A., Breininger, K., et~al.: Pan-tumor canine
  cutaneous cancer histology ({CATCH}) dataset. Scientific Data  \textbf{9}(1),
   1--13 (2022)

\bibitem{xu2022clinical}
Xu, Z., Myronenko, A., Yang, D., Roth, H.R., Zhao, C., Wang, X., Xu, D.:
  Clinical-realistic annotation for histopathology images with probabilistic
  semi-supervision: A worst-case study. In: Medical Image Computing and
  Computer Assisted Intervention--MICCAI 2022: 25th International Conference,
  Singapore, September 18--22, 2022, Proceedings, Part II. pp. 77--87. Springer
  (2022)

\bibitem{yang2017suggestive}
Yang, L., Zhang, Y., Chen, J., Zhang, S., Chen, D.Z.: Suggestive annotation: A
  deep active learning framework for biomedical image segmentation. In:
  International conference on medical image computing and computer-assisted
  intervention. pp. 399--407. Springer (2017)

\end{thebibliography}

\newpage 
\setcounter{page}{1}
\setcounter{section}{0}

\title{Adaptive Region Selection for Active Learning in Whole Slide Image Semantic Segmentation}
% with increased robustness}
%
\titlerunning{Adaptive Region Selection for AL in WSI Semantic Segmentation}
% \author{Anonymous}
% \institute{Anonymous Organization\\
% \email{***@***.***}}
\author{Jingna Qiu\inst{1} \and 
Frauke Wilm\inst{1,2} \and 
Mathias Öttl\inst{2} \and
Maja Schlereth\inst{1} \and
Chang Liu\inst{2} \and
Tobias Heimann\inst{3} \and
Marc Aubreville\inst{4} \and
Katharina Breininger\inst{1}}
% index{Qiu, Jingna, Wilm, Frauke, Öttl, Mathias, Schlereth, Maja, Liu, Chang, Heimann, Tobias, Aubreville, Marc, Breininger, Katharina}
\authorrunning{Qiu et al.}
\institute{Department Artificial Intelligence in Biomedical Engineering, Friedrich-Alexander-Universität Erlangen-Nürnberg, Erlangen, Germany \\
\email{jingna.qiu@fau.de} \and
Pattern Recognition Lab, Department of Computer Science, \\ Friedrich-Alexander-Universität Erlangen-Nürnberg, Erlangen, Germany \and
Digital Technology and Innovation, Siemens Healthineers, Erlangen, Germany \and
Technische Hochschule Ingolstadt, Ingolstadt, Germany\\}
\maketitle              % typeset the header of the contribution
\section{Supplementary Materials}
\supplementaryfigures
\begin{figure}
    \centering
    \includegraphics[width=.6\textwidth]{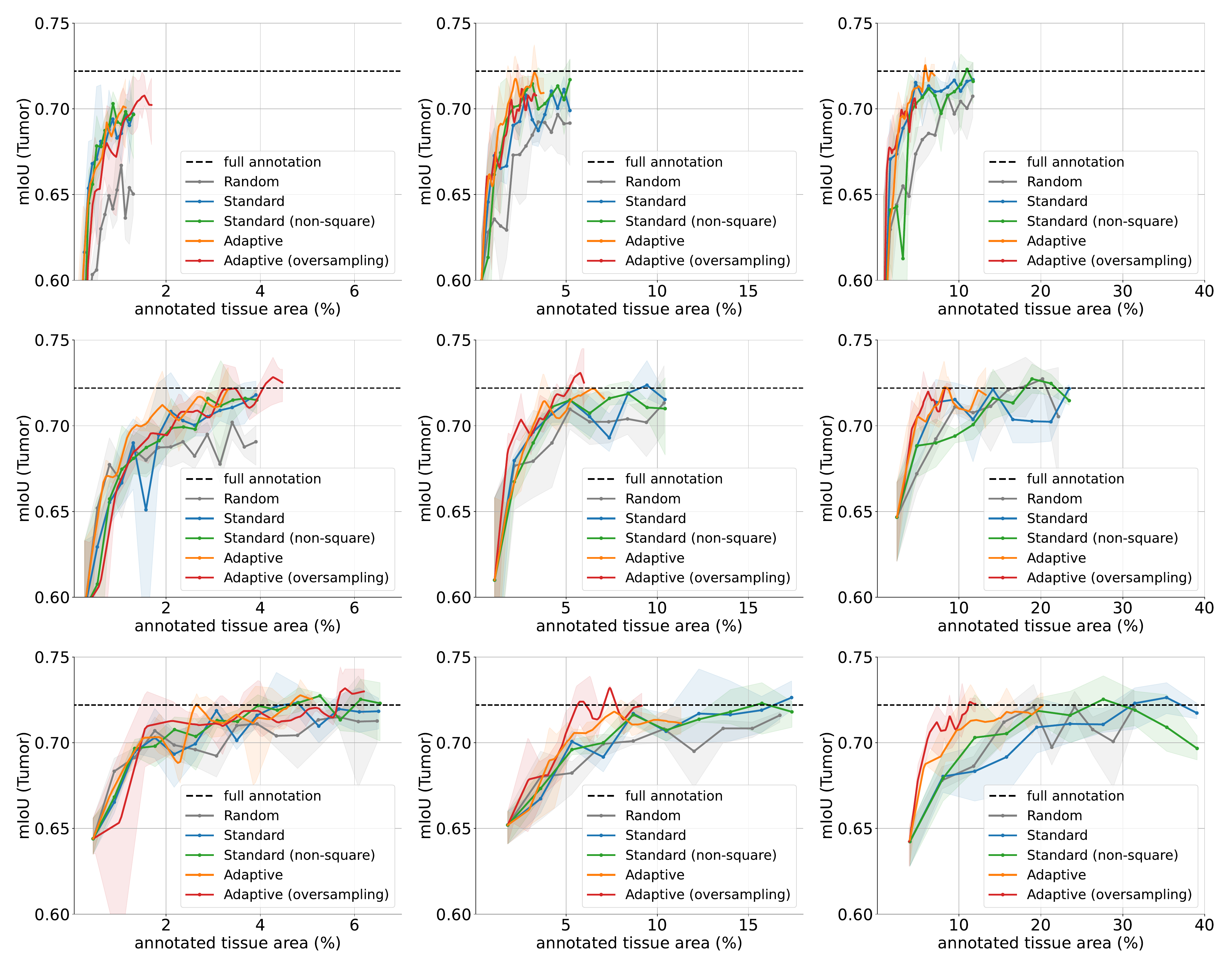}
    \caption{Supplementary results for Fig.~\ref{fig:step_size_result} (h). Allowing for selection of previously annotated regions improves the sampling efficiency of \textit{Adaptive}, thereby facilitating the attainment of full annotation performance. Note that these results are from a preliminary study, and a systematic evaluation of sampling strategies within the \textit{annotated} dataset (e.g., oversampling of annotated regions that remain uncertain) is subject to future work.}
    \label{fig:step_size_result_oversample}
\end{figure}
\end{document}